\documentclass{article}

\usepackage[final]{corl_2026} 

\usepackage{amsmath}
\usepackage{amsfonts}
\usepackage{algorithm}
\usepackage{algpseudocode}
\usepackage{booktabs}
\usepackage{graphicx}
\usepackage{enumitem}
\usepackage{amssymb}
\usepackage{cleveref}
\usepackage{float}
\usepackage{placeins}
\usepackage{wrapfig}
\usepackage{pifont}
\DeclareUnicodeCharacter{2460}{\ding{172}}
\DeclareUnicodeCharacter{2461}{\ding{173}}
\DeclareUnicodeCharacter{2462}{\ding{174}}
\newcommand{\samplecount}[1]{(\#\,#1)}

\title{Never Too Late for Force: Accelerating VLA Post-Training with Reactive Force Injection}

\author{
  \parbox[t]{\dimexpr\textwidth-2\tabcolsep-2.5pt\relax}{\centering\normalfont
    Yi Wang$^{12*}$, Wendi Chen$^{12*\ddagger}$, Zimo Wen$^{14*}$ \\
    Han Xue$^{1}$, Xueqi Li$^{23}$, Wenye Yu$^{12}$, Zhijie Chen$^{1}$, Hao Yang$^{1}$, Jun Lv$^{5}$ \\
    Chuan Wen$^{1\dagger}$, Cewu Lu$^{125\dagger}$ \\
    {\normalsize $^{1}$Shanghai Jiao Tong University \enspace $^{2}$Shanghai Innovation Institute} \\
    {\normalsize $^{3}$Southern University of Science and Technology} \\
    {\normalsize $^{4}$Zhiyuan College, Shanghai Jiao Tong University \enspace $^{5}$Noematrix Ltd.} \\
    \small $^{*}$Equal contribution \quad $^{\ddagger}$Project lead \quad $^{\dagger}$Corresponding authors \\
    \small \href{https://lift-policy.github.io}{lift-policy.github.io}
  }
}

\begin{document}
\maketitle


\begingroup
\setlength{\intextsep}{4pt}
\begin{figure}[H]
  \centering
  \vspace{-22pt}
  \makebox[\linewidth][c]{\includegraphics[width=1.08\linewidth]{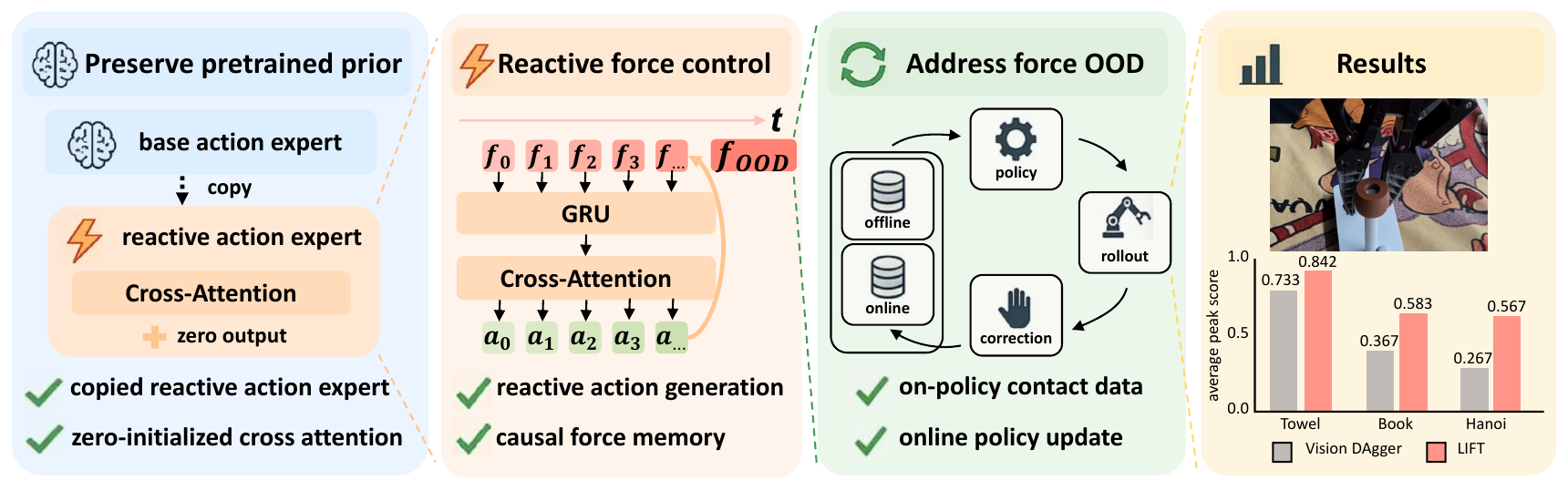}}
  \vspace{-8pt}
  \caption{\textbf{LIFT overview.}
  (Left) LIFT copies the pretrained base action expert to initialize a reactive action expert, with zero-initialized force-injected cross attention preserving the original policy's output at initialization.
  (Center-left) Causal force memory encodes recent force history and injects it through cross attention for reactive within-chunk action updates.
  (Center-right) To address force distribution shift (force OOD), online DAgger collects on-policy contact corrections and updates the policy using mixed offline and online data.
  (Right) LIFT achieves higher peak and final performance than vision-only DAgger across the three evaluated manipulation tasks.}
  \label{fig:teaser}
\end{figure}
\endgroup

\begin{abstract}
Pretrained vision-language-action (VLA) policies provide strong language-conditioned manipulation knowledge, but they remain largely vision-driven and can struggle once manipulation enters contact states where the scene is occluded, depth is ambiguous, or small force errors push execution off the offline demonstration distribution. We present \textbf{LIFT}, \emph{\textbf{L}ate Reactive \textbf{I}njection of \textbf{F}orce for VLA Post-\textbf{T}raining}, a force-aware post-training framework that adds contact reactivity to a pretrained VLA policy while preserving its output at initialization. LIFT grafts a reactive action expert beside the original action expert, initializes it from pretrained action weights, and injects recent 6D end-effector force through \emph{causal force memory} and \emph{zero-initialized cross attention}, enabling actions to be refreshed during execution. To address the policy-dependent distribution shift of contact feedback, LIFT further couples reactive force injection with an online DAgger loop that trains on a mixture of offline task-alignment data and human-corrected online rollouts. Across towel folding, book insertion, and Hanoi ring placement, LIFT learns faster and reaches higher peak and final performance than vision-only DAgger, while ablations show that reactive force memory and online corrective data are both important for robust contact-rich manipulation. Our code is publicly available at \url{https://github.com/y-wng/lift}.
\end{abstract}



\section{Introduction}

Pretrained vision-language-action (VLA) policies \citep{brohan2023rt2,octo2024,kim2024openvla,black2024pi0,intelligence2025pi05,generalist2026gen1} have made it possible to transfer large-scale vision-language priors to robot manipulation. These models are strong at language-conditioned scene understanding and general action generation, but they still rely mainly on vision. Without force sensing, a policy can struggle when contact states are visually ambiguous due to occlusion, partial observation, or inaccurate depth perception. This limitation matters even when a task is not explicitly contact-rich, because execution still passes through contact states where physical feedback can reveal whether the robot has touched, pushed, inserted, or released an object. In those moments, force provides extra physical feedback and can also serve as a cheap memory of contact history. Recent force-aware manipulation systems \citep{he2024foar,liu2024forcemimic,hou2024acp,xue2025rdp,chen2025implicitrdp,li2025vtam,wang2025ftp1} further support this view by showing that explicit contact feedback helps in both reactive control and policy learning. However, force and torque are harder to include during pretraining at scale because they are costly to collect, depend on robot platform and end-effector hardware, and vary across setups.

This makes force-aware post-training a natural next step. Instead of rebuilding the foundation model, we want to adapt an existing VLA to a specific robot platform and sensor stack. Doing so, however, raises three questions. First, because force is high-frequency, low-dimensional, and tightly coupled to contact dynamics, how should it be injected so the policy can react quickly to new contact events and retain a useful force memory? Second, adding a new force path can disturb the pretrained VLA prior, so how can we preserve it at initialization and during adaptation? Third, force distributions are highly policy-dependent and can shift sharply after small deviations from offline states, so how can post-training stay effective when offline force data do not cover most states of force?

We address these questions with \textbf{LIFT}, \emph{\textbf{L}ate Reactive \textbf{I}njection of \textbf{F}orce for VLA Post-\textbf{T}raining}, a force-aware post-training framework for pretrained VLA models. As summarized in \Cref{fig:teaser}, LIFT adds a reactive action expert beside the original action expert and injects \emph{causal force memory} through a causal force encoder and cached vision-language prefix. This gives the policy a fast reactive path that can update actions in real time while treating force as a low-cost temporal context over recent contact history. To avoid damaging the pretrained model, LIFT copies the original action parameters into the reactive action expert, uses \emph{shifted causal attention}, and adds \emph{zero-initialized cross attention} so the network starts output-equivalent to the original VLA. To handle noisy and shifting force distributions, LIFT couples this architecture with online DAgger, so the model can keep collecting corrections from the states induced by its own policy and adapt to the resulting contact distribution.

We evaluate LIFT on three real-world manipulation tasks with different contact characteristics: towel folding, book insertion, and Hanoi ring placement. Across these tasks, LIFT improves faster than vision-only DAgger and reaches higher peak and final performance. Under the shifts of objects, tablecloths, and lighting, LIFT shows no substantial degradation relative to its in-distribution behavior. Online post-training consistently outperforms offline-only adaptation on all three tasks.

Our contributions are listed below.
\begin{itemize}[leftmargin=2.0em,labelsep=0.5em]
    \item We design reactive force injection with \emph{causal force memory} and cached vision-language prefix so VLA can react to contact changes while using force as a low-cost temporal context.
    \item We preserve the pretrained VLA prior with initialization-equivalent parameter copying, \emph{shifted causal attention}, and \emph{zero-initialized cross attention}.
    \item We propose a force-aware online post-training framework that adapts pretrained VLA models to platform-specific force feedback and mitigates force distribution shift.
\end{itemize}

\section{Related Work}
\label{sec:related}

\paragraph{Force-augmented policies.}
Force-aware manipulation uses physical feedback to resolve contact states that are difficult to infer from images alone. Prior approaches include visual-force goals~\citep{collins2023forcesight}, force-gated diffusion~\citep{he2024foar}, hybrid force-position imitation~\citep{liu2024forcemimic}, and learned compliance~\citep{hou2024acp}. Other methods use force-attending curricula~\citep{liu2025factr}, force-informed kinesthetic demonstrations~\citep{chen2025dexforce}, or interaction-frame control~\citep{fang2026forcepolicy}. Reactive policies also address the mismatch between visual inference speed and contact dynamics. RDP~\citep{xue2025rdp} couples slow visual diffusion with a fast tactile-conditioned decoder. ImplicitRDP~\citep{chen2025implicitrdp} unifies visual-force diffusion with GRU memory and causal attention, allowing new force observations to update actions within a chunk. FA-RDP~\citep{zhuo2026fardp} uses a learned multimodality indicator to adapt inference frequency and diffusion sampling steps, balancing diverse pre-contact motions with fast contact responses. LIFT builds on these reactive mechanisms to turn a non-reactive, MOT-based policy into a reactive one that can fully utilize force information.

\paragraph{Vision-language-action models for manipulation.}
General VLAs~\citep{brohan2023rt2,octo2024,kim2024openvla,generalist2026gen1}, including $\pi_0$/$\pi_{0.5}$~\citep{black2024pi0,intelligence2025pi05}, learn primarily from visual observations and robot trajectories. Collecting synchronized force or torque data at comparable scale is costly, motivating the addition of these modalities during post-training. ForceVLA~\citep{yu2025forcevla} integrates force through expert routing in the action decoder. TA-VLA~\citep{zhang2025tavla} uses decoder-side torque adapters, compact history summaries, and auxiliary torque prediction. ForceVLA2~\citep{li2026forcevla2} combines force prompts and multi-scale fusion with hybrid force-position outputs. While these approaches develop mechanisms for fusing contact feedback into action prediction, FAVLA~\citep{li2026favla} additionally connects force conditioning to execution timing: it couples a slow vision-language model with a fast force-conditioned expert and uses predicted force variation to schedule its updates. LIFT introduces force fusion and reactive execution into pretrained policies while preserving their initial outputs to retain prior capabilities and knowledge as much as possible.

\paragraph{Post-training for robotic policies.}
Small contact errors can move a robot into states absent from its offline demonstrations. DAgger~\citep{ross2011dagger} and interactive variants HG-DAgger~\citep{kelly2019hgdagger}, SIRIUS~\citep{liu2022sirius}, and Sirius-Fleet~\citep{liu2024siriusfleet} address this distribution shift through expert corrections on learner-visited states. SOE~\citep{jin2025soe} explores policy self-improvement through on-manifold exploration. Recent infrastructure supports online post-training~\citep{pan2026sop,fang2026robopocket}. CR-DAgger~\citep{xu2025crdagger} is especially relevant to contact-rich tasks: compliant human corrections train a residual policy that outputs both position and force adjustments without interrupting the base policy. LIFT incorporates force information into a pretrained policy through post-training with online corrections.

\section{Problem Formulation}
  \label{sec:problem}

  Starting from a vision-only
  VLA policy, our ultimate goal is to introduce force late in post-training without discarding its
  general manipulation knowledge. To achieve this, we use a two-stage training pipeline. In
  Stage~1, a handheld data-collection device provides an abundant vision-only task-alignment dataset
  $D_v=\{(\ell,I,\mathbf{a})\}$, from which a vision-only policy $\pi^{\mathrm{V}}(\mathbf{a}\mid \ell,I)$ learns coarse
  manipulation knowledge. In
  Stage~2, we switch to a real robot equipped with a 6D end-effector force sensor and collect online corrective data
  $D_f^{(k)}=\{(\ell,I,\mathbf{F},\mathbf{a}^{*})\}$ (where $\mathbf{a}$ and $\mathbf{F}$ are chunks from $t$ to $t+H-1$). The post-training policy
  $\pi^{\mathrm{V+F}}(\mathbf{a}\mid \ell,I,\mathbf{F})$ is trained on the online sequence $\{D_f^{(k)}\}_k$ together with the
  static visual set $D_v$. Because the second-stage data
  are collected by online DAgger from the on-policy distribution $d_{\pi^{(k)}}$, the corrective set is effectively reduced to
  the states actually visited by the current policy, which helps mitigate covariate shift. Our hypothesis is that this force participation should speed up
  post-training rather than merely add another modality.

  To realize this overall goal, we must solve three technical objectives:
  \begin{itemize}[leftmargin=2.0em,labelsep=0.5em]
    \item \textbf{Reactive force injection (O1).} Inject recent force memory into the policy so it can react to contact events
    instead of committing to an open-loop action chunk.
    \item \textbf{Preservation of general capabilities (O2).} Preserve the model's generalization ability while still improving
    its ability to adjust actions promptly during contact.
    \item \textbf{Simultaneous use of heterogeneous data (O3).} Jointly exploit abundant visual task-alignment data and scarce
    force-enabled online corrections so the policy benefits from both supervision sources throughout post-training.
  \end{itemize}

\section{Methods}
\label{sec:method}

Our methods are designed around the three objectives in \Cref{sec:problem}. To satisfy O1, we add a late reactive force injection path that can causally decode actions within a chunk from recent force measurements while reusing the cached vision-language prefix. To satisfy O2, we initialize this new path so that the augmented model is output-equivalent to the original $\pi_{0.5}$ before post-training. To satisfy O3, we train the same model on both vision-only offline data and force-enabled online corrections through explicit force masking and balanced sampling.

\subsection{Closing the loop with reactive force injection (O1)}
\label{sec:method-reactive}

\begin{figure}[htbp]
  \centering
  \includegraphics[width=0.8\linewidth]{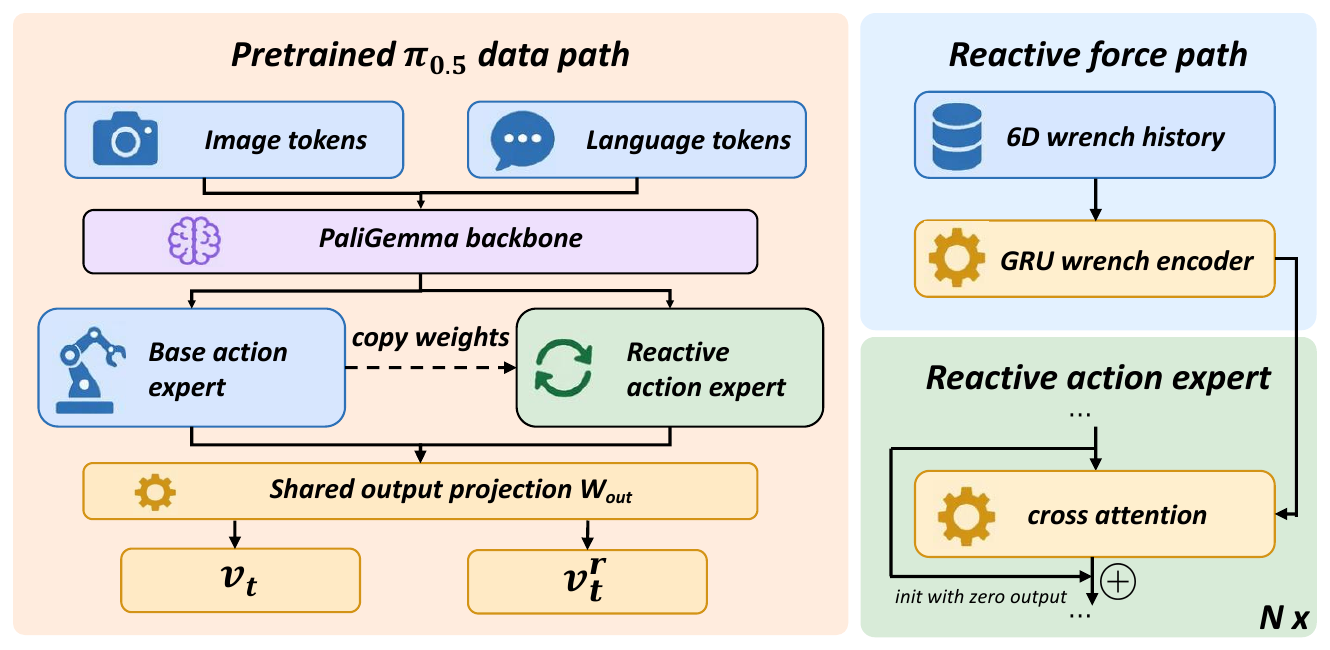}
  \vspace{-4pt}
  \caption{\textbf{LIFT architecture.}
  LIFT grafts a reactive action expert beside pretrained $\pi_{0.5}$ while keeping the original vision-language stack unchanged. Recent force is encoded as force memory and injected into the reactive branch through zero-initialized cross attention, enabling within-chunk action updates while preserving the pretrained prior at initialization.}
  \label{fig:arch}
\end{figure}
\Cref{fig:arch} summarizes the O1 mechanism. A pretrained $\pi_{0.5}$ action expert normally denoises a whole action chunk with full within-chunk attention, which is not suitable for updating individual actions from newly observed contact. LIFT first turns the action side into a causal reactive stream that decodes the chunk action by action, then injects latency-aligned force memory into that stream, and caches the slow vision-language context so the reactive path can be refreshed within each chunk.

\paragraph{O1.1 Reactive action expert.}
We instantiate a reactive action expert beside the original $\pi_{0.5}$ action expert and use it as the stream that will be updated during contact. For O1, the key requirement is that action decoding in this new expert is causal within each chunk. This makes the reactive stream suitable for refreshing the chunk action by action as new contact arrives. The exact shifted causal attention pattern and token-level formulation are deferred to O2.1 and highlighted in \Cref{fig:mask}.

\paragraph{O1.2 Causal force-injected cross attention.}
LIFT encodes recent 6D end-effector force into a causal force memory and injects it only into the reactive action expert through force-injected cross attention with a latency-aligned causal mask. This lets each reactive action use the newest force measurements available when inference finishes, so the controller can react to contact in real time without using outdated or unavailable force, with the full formulation deferred to \Cref{app:mask-details}.

\paragraph{O1.3 Cached Vision-Language Slow Context.}
The final obstacle to closed-loop contact response is latency from the vision-language context. LIFT separates this slow context from the fast force-injected action update. At deployment, the runtime first computes the vision-language prefix once and stores its KV cache. Within the action chunk, it then repeatedly encodes the latest latency-aligned force history and reevaluates the action experts against the cached prefix. Because the expensive vision-language prefix is reused, each refresh can update the reactive action chunk with newly observed contact transients without waiting for a full vision-language forward pass.

\subsection{Preserving the pretrained VLA prior at initialization (O2)}
\label{sec:method-preserve}

Reactive force injection gives LIFT the closed-loop path required by O1, but it also alters the pretrained VLA's action-side structure. Such structural change can make the model forget the original manipulation capabilities that make the pretrained prior useful. We therefore preserve the prior at initialization through an output-equivalent reactive action expert and zero-initialized cross attention.

\begin{wrapfigure}{r}{0.49\textwidth}
  \centering
  \includegraphics[width=0.47\textwidth,trim=0 16pt 0 16pt,clip]{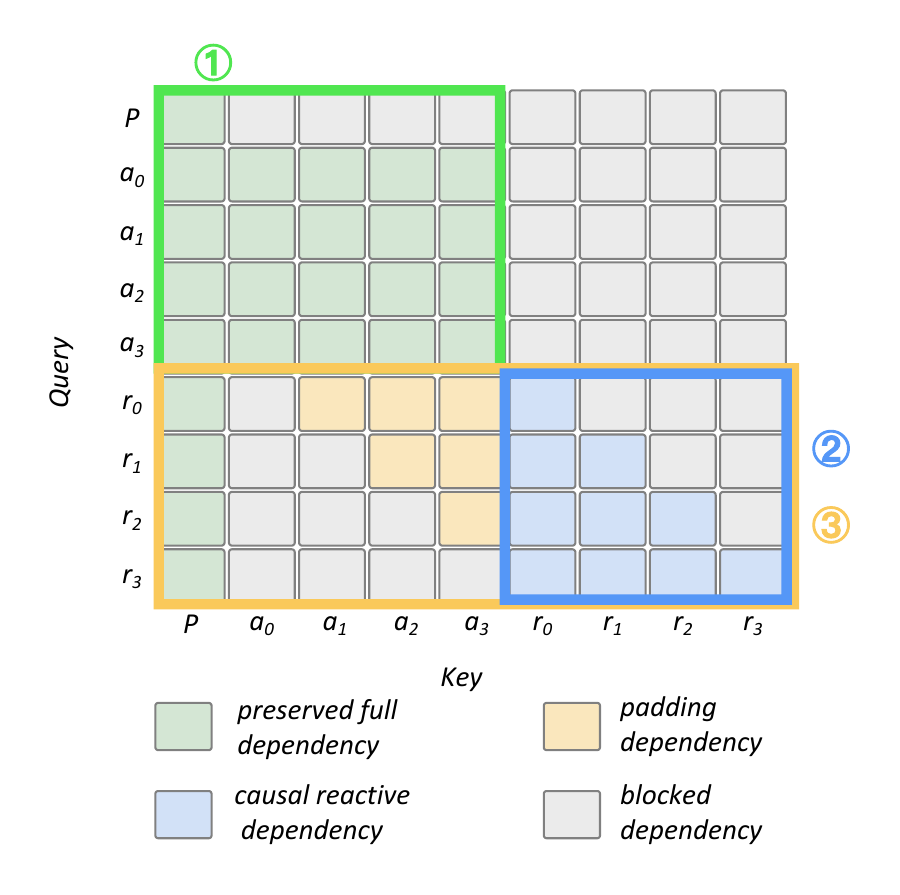}
  \vspace{-2pt}
  \caption{\textbf{Shifted causal attention with initialization equivalence.}
  Each reactive token attends to the vision-language prefix, later base-action tokens, and the causal reactive prefix. Combined with the copied pretrained action expert, LIFT preserves initialization equivalence while enabling causal reactive updates.}
  \label{fig:mask}
\end{wrapfigure}
\paragraph{O2.1 Make reactive action expert output-equivalent.}
Starting from pretrained $\pi_{0.5}$, we instantiate a second action-denoising expert with the same action-token interface as the original expert and copy the original action-expert weights into it. The original action expert is fully attentive within a chunk: each base action token $a_i$ can attend to all action tokens $a_{0:H-1}$. LIFT keeps this base action stream unchanged, as shown by ① in \Cref{fig:mask}, then appends reactive tokens $r_{0:H-1}$ with the shifted causal attention introduced in O1.1 and illustrated in ③ of \Cref{fig:mask}. For each position $i$, the reactive token $r_i$ sees the vision-language prefix, later base-action tokens $a_{i+1:H-1}$, and the causal reactive prefix $r_{0:i}$, while blocking base-action tokens $a_{0:i}$ and subsequent reactive tokens $r_{i+1:H-1}$. At initialization, copied weights and aligned positions make $r_{0:i}$ equivalent to base-action representations $a_{0:i}$, while the shifted context supplies $a_{i+1:H-1}$ from the base stream. Thus $r_i$ receives the same context as $a_i$ in the original fully-attentive action expert, making ① and ③ in \Cref{fig:mask} initialization-equivalent and the reactive action expert output-equivalent to the original before force training.

\paragraph{O2.2 Zero-initialize cross attention output.}
The zero-initialized cross attention block is added as a residual update to the reactive action representation. LIFT zero-initializes the output projection of this residual path, so the force-injected update is exactly zero at step~0 even though the force encoder and attention logits are present. With no effective force residual, the copied reactive action expert receives the same action interface and output projection as the original action expert. Thus force cannot change the pretrained action output until post-training learns a nonzero residual.

Together, O2.1 and O2.2 let LIFT add the causal action branch and force path needed for O1 while starting post-training from the same action output as the base model.

\subsection{Training with heterogeneous visual and force data (O3)}
\label{sec:method-data}

O3 turns the architecture above into a post-training recipe that uses both available supervision sources. The offline set $D_v$ provides broad vision-only task alignment from the handheld data-collection device, while the online set $D_f$ provides force-enabled corrections on states visited by the deployed policy. LIFT trains both streams with one shared objective, but uses explicit force masking so samples without force data do not update the force encoder.

\paragraph{O3.1 Additive flow-matching objective.}
For both data sources, LIFT jointly trains the original and reactive action streams. We sample independent noises $\varepsilon,\varepsilon^{\text{r}}\sim\mathcal{N}(0,I)$ and same-time interpolants $x_\tau,x_\tau^{\text{r}}$, then minimize
\begin{equation}
\mathcal{L}(\theta) \;=\; \big\lVert v_\theta(x_\tau,\tau,o) - u \big\rVert_2^2 \;+\; \big\lVert v_\theta^{\text{r}}(x_\tau^{\text{r}},\tau,o,m_{t:t+H}) - u^{\text{r}} \big\rVert_2^2 .
\label{eq:loss}
\end{equation}
The two losses are computed in the same forward pass, their gradients are accumulated together, and the shared parameters are updated jointly. This keeps the base stream anchored to the pretrained action prior while the reactive stream learns force-injected corrections. At inference time, LIFT still computes both action streams because the reactive stream uses the base stream through shifted causal attention, but only the reactive action is sent to the robot controller.

\paragraph{O3.2 Selective force masking.}
Each sample enables the force path only when real force data exists. Vision-only batches use zero force placeholders for shape, then mask encoded force memory before cross attention, blocking gradients to the force encoder and force-injected attention. Online correction batches keep measured force memory active, updating the force pathway and reactive branch. Thus one model trains on both data sources without synthetic force labels.

\paragraph{O3.3 Equal sampling of heterogeneous data.}
We follow the RLPD symmetric sampling strategy~\citep{ball2023efficient} and train on a $1{:}1$ mixture of offline task-alignment batches and online corrective batches. Each update therefore sees equal amounts of prior-preserving vision-only data and force-enabled on-policy corrections, allowing the model to benefit from the complementary strengths of both sources.

\subsection{Pipeline overview and system implementation}
\label{sec:method-pipeline}

\begin{figure}[htbp]
  \centering
  \includegraphics[width=0.82\linewidth]{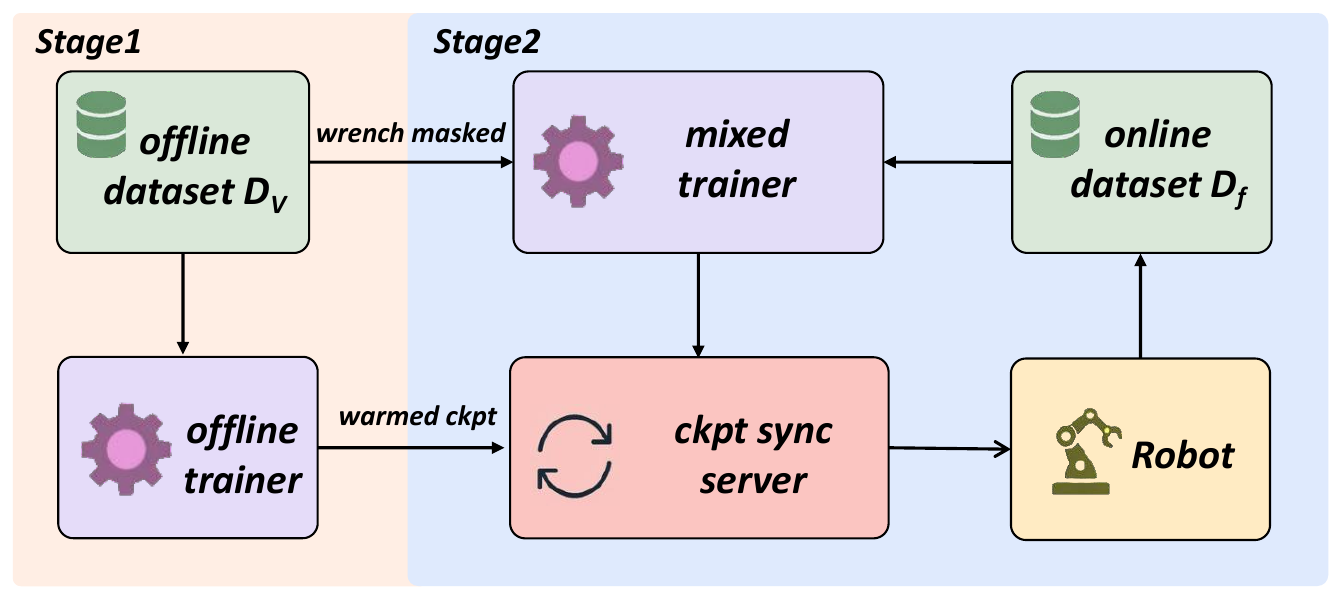}
  \vspace{-4pt}
  \caption{\textbf{LIFT training pipeline.}
  The two-stage system first aligns LIFT on vision-only demonstrations and then closes an online post-training loop between the robot and cloud trainer.}
  \label{fig:system}
\end{figure}

As shown in \Cref{fig:system}, the pipeline instantiates the three objectives in two stages. Stage~1 builds LIFT on pretrained $\pi_{0.5}$ and trains on the vision-only dataset $D_v$ with force masked out, so the model aligns to the target tasks while preserving the pretrained prior. Stage~2 moves to the real robot, collects force-enabled corrective episodes into $D_f$, and continues post-training on a fixed $1{:}1$ mixture of $D_v$ and the asynchronously ingested online buffer $D_f$. Similar to recent work~\citep{pan2026sop,fang2026robopocket} on online post-training infrastructure, the latest checkpoint is periodically synced back to the rollout client for redeployment, closing the loop between correction collection, post-training, and redeployment.


\section{Experimental Results}
\label{sec:result}

\subsection{Experimental setup}
We evaluate on three real-robot manipulation tasks: towel folding, book insertion, and Hanoi ring placement. The stage-one vision-only data are collected with a handheld data-collection device, while stage-two force-enabled post-training data are collected through Flexiv TDK~\citep{flexivtdk2026}. Rollouts are executed on a Flexiv Rizon~4S robot with a 6D end-effector force sensor. Each checkpoint is evaluated with $3 \times 10$ autonomous rollouts (three groups of ten), from which we compute 95\% confidence intervals. Collected samples (\#) count human-intervention frames from online collection, which are the only online data used for training. Towel and book use stage-based scores, while Hanoi uses binary success. Scoring, task protocols, and failure case analysis are detailed in \Cref{app:tasks}.

We compare six policies:
\begin{itemize}[leftmargin=2.0em,labelsep=0.5em]
  \item {\boldmath\textbf{$\pi_{0.5}$ w/ Online DAgger}}: the same online DAgger loop without force input.
  \item \textbf{LIFT w/o Reactive Force Injection}: stage-two force post-training with a single-frame force input rather than the reactive force memory.
  \item \textbf{LIFT w/ Offline DAgger}: the reactive model trained with a fixed force-correction buffer, without repeated online updates.
  \item {\boldmath\textbf{$\pi_{0.5}$ w/ Offline Handheld Data}}: a policy trained only on the larger offline handheld set.
  \item \textbf{Residual Policy}: a lightweight force-reactive Transformer that adds action corrections to a frozen, offline-trained $\pi_{0.5}$ policy.
  \item \textbf{LIFT}: the full reactive force-injected VLA trained with online DAgger.
\end{itemize}

\subsection{Does force accelerate or improve post-training? (Q1)}
\label{sec:q-speed}
\begin{figure*}[t]
  \centering
  \includegraphics[width=\linewidth,trim={0 6pt 0 0},clip]{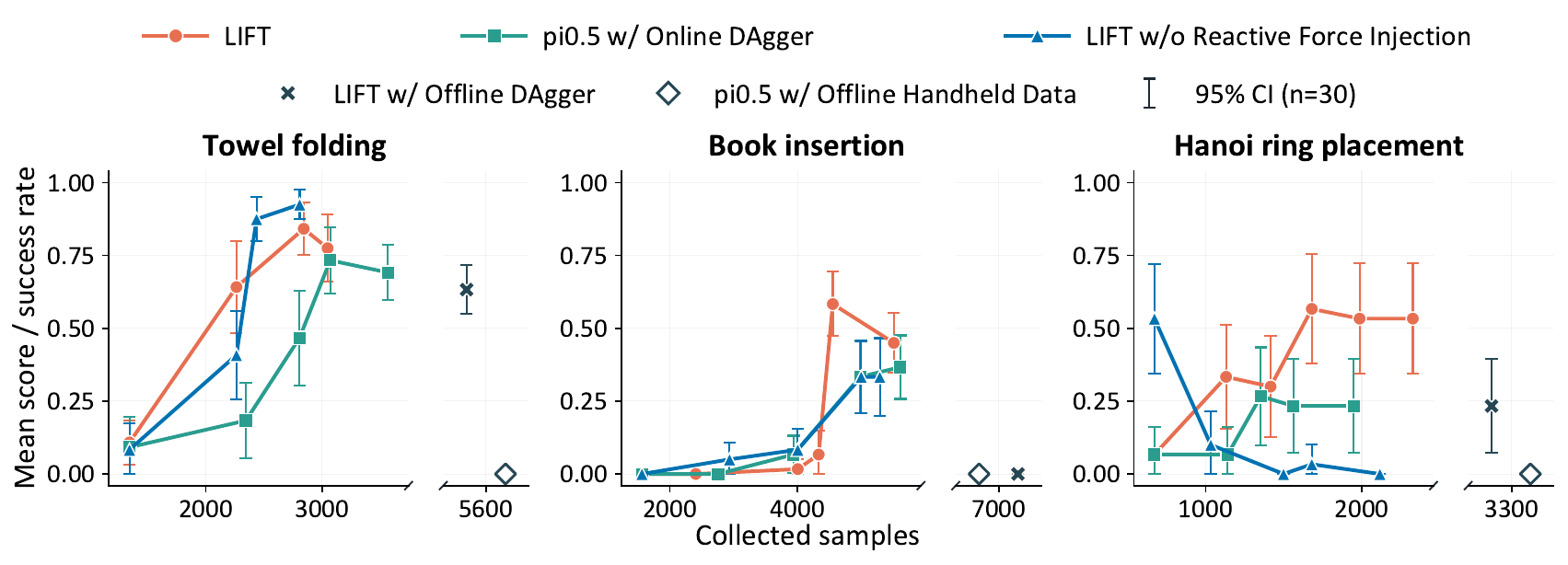}
  \vspace{-1.1em}
  \caption{\textbf{Performance curves on three tasks as sample count increases.}
  By leveraging reactive control and force memory, LIFT reaches higher peak performance with fewer samples.}
  \label{fig:online-curves}
\end{figure*}

\Cref{fig:online-curves} shows that force accelerates post-training and raises peak performance across contact-rich manipulation tasks. The corresponding optimizer-step curves, online experiment duration, and episode counts are provided in \Cref{app:steps-curves}.

On towel folding, both force-aware variants improve faster and reach higher scores than vision-only online DAgger. $\pi_{0.5}$ w/ Online DAgger improves from $0$ \samplecount{1.3K} to $0.725$ \samplecount{3.1K}, while LIFT reaches $0.65$ with fewer samples \samplecount{2.3K} and peaks at $0.825$ \samplecount{2.8K}. LIFT w/o Reactive Force Injection improves even faster on this task, reaching $0.925$ \samplecount{2.4K} and peaking at $0.95$ \samplecount{2.8K}. The main failure case is contact ambiguity: because the towel is thin, a monocular wrist camera can misestimate depth and fail to tell whether the gripper has actually contacted or grasped the cloth. Force feedback directly exposes this contact state, helping the policy recover from failed grasps during the first and subsequent folds.

Book insertion highlights the value of the full LIFT policy for constrained insertion. LIFT reaches the highest peak score, $0.6$ \samplecount{4.6K}, compared with $0.4$ \samplecount{5.6K} for $\pi_{0.5}$ w/ Online DAgger. The task requires inserting a book into an empty slot bounded by neighboring books and the shelf, so side contact and insertion depth determine when the robot should keep pushing, stop, or release. The key vision-only failure is that the policy cannot reliably judge whether the book has bottomed out: it may continue pushing after full insertion or release before the book is seated, causing the book to flip or fail to settle. LIFT uses contact feedback from the book--shelf interaction to learn the insertion and release timing needed for higher final performance.

Hanoi ring placement shows the same pattern in a precision placement setting. LIFT reaches $0.6$ \samplecount{1.7K} and maintains $0.6$ at the final checkpoint \samplecount{2.0K}, whereas $\pi_{0.5}$ w/ Online DAgger peaks at $0.3$ \samplecount{1.4K} and ends at $0.2$ \samplecount{1.6K}. Here the clearance between the ring and pole is small, so slight misalignment can turn a placement attempt into a jam. The core vision-only failure is continuing to push downward when the ring is not fully aligned, which tilts the ring and prevents insertion. By sensing the direction and magnitude of contact forces, LIFT can identify offset or stuck states and adjust the motion, making force feedback useful for both faster learning and higher scores.

\FloatBarrier
\subsection{Does reactivity matter beyond single-frame force? (Q2)}
\label{sec:q-history}
We compare LIFT with LIFT w/o Reactive Force Injection to study whether force should enter as reactive force memory rather than as a single-frame cue.

On book insertion, LIFT reaches a peak score of $0.6$ \samplecount{4.6K}, while the single-frame force baseline peaks at $0.4$ \samplecount{5.0K} and \samplecount{5.3K}. A typical failure of LIFT w/o Reactive Force Injection occurs after the book has already bottomed out in the slot: the policy keeps pushing downward against the shelf, even under large contact force, instead of releasing the gripper or backing out. This non-Markovian decision requires knowing whether the current load comes from ongoing insertion or from a completed insertion that should trigger release. LIFT uses recent force memory to infer the contact phase from how the force evolved, so it can stop pushing and complete the final insertion-and-release stage more reliably.

On Hanoi ring placement, LIFT reaches a peak score of $0.6$ \samplecount{1.7K}, while LIFT w/o Reactive Force Injection peaks at $0.5$ \samplecount{0.7K} and later drops to $0.1$ \samplecount{1.0K}, $0$ \samplecount{1.5K}, and $0.1$ \samplecount{1.7K}. The typical single-frame failure appears when the ring collides with the pole or surrounding support: an abnormal force spike can drive the policy into unstable motions, including oscillating forward and backward, losing localization, or moving far away from the placement region. Fast and precise correction is required because contact can occur around the whole ring. Single-frame force is easily misled by instantaneous noise or out-of-distribution impact forces. LIFT makes the policy reactive by averaging this decision over a short force memory window and injecting the contact signal into the reactive branch, which stabilizes the force input and lets the policy update its action chunk quickly after contact changes.

On towel folding, single-frame force achieves performance comparable to the reactive variant, as the task involves relatively simple contact states.

\subsection{Does LIFT preserve original VLA generalization? (Q3)}
\label{sec:q-shifts}
\begin{figure}[htbp]
  \centering
  \includegraphics[width=\linewidth]{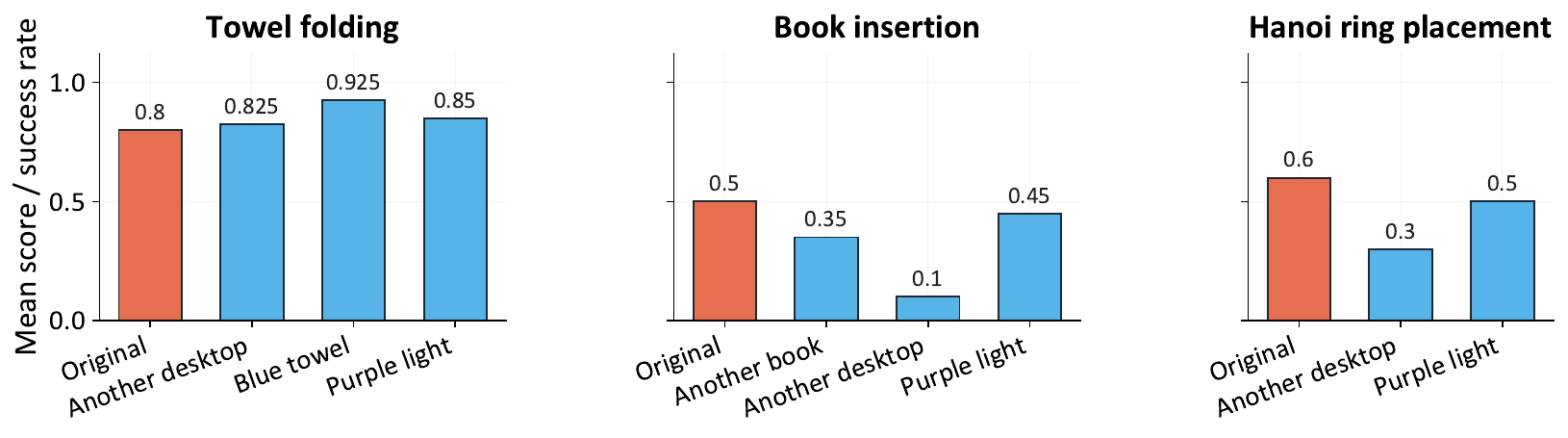}
  \vspace{-4pt}
  \caption{\textbf{Generalization of the final LIFT checkpoint.} LIFT shows no substantial degradation under shifts of objects, tablecloths, and lighting.}
  \label{fig:reactive-generalization}
\end{figure}

Under shifts of objects, tablecloths, and lighting, \Cref{fig:reactive-generalization} shows no substantial degradation for LIFT on most tested settings across the three tasks. The exact shift introduced in each condition (e.g., \emph{Another desktop}, \emph{Blue towel}, and \emph{Purple light}) is documented in \Cref{app:gen}. This suggests that LIFT largely preserves original VLA generalization, consistent with the copied action expert, shifted causal attention, and zero-initialized cross attention.

\subsection{Does online data matter for reactive force injection? (Q4)}
\label{sec:q-online}
Offline-only comparisons are plotted as separate markers in \Cref{fig:online-curves}. LIFT w/ Offline DAgger underperforms on all three tasks and falls to zero on book insertion. This indicates that offline data miss abrupt force-pattern changes during real execution, while online DAgger adds corrections on the learner's failure states.

\subsection{Why use LIFT rather than a lightweight residual policy? (Q5)}
\label{sec:q-residual}
\begin{figure}[htbp]
  \centering
  \includegraphics[width=\linewidth]{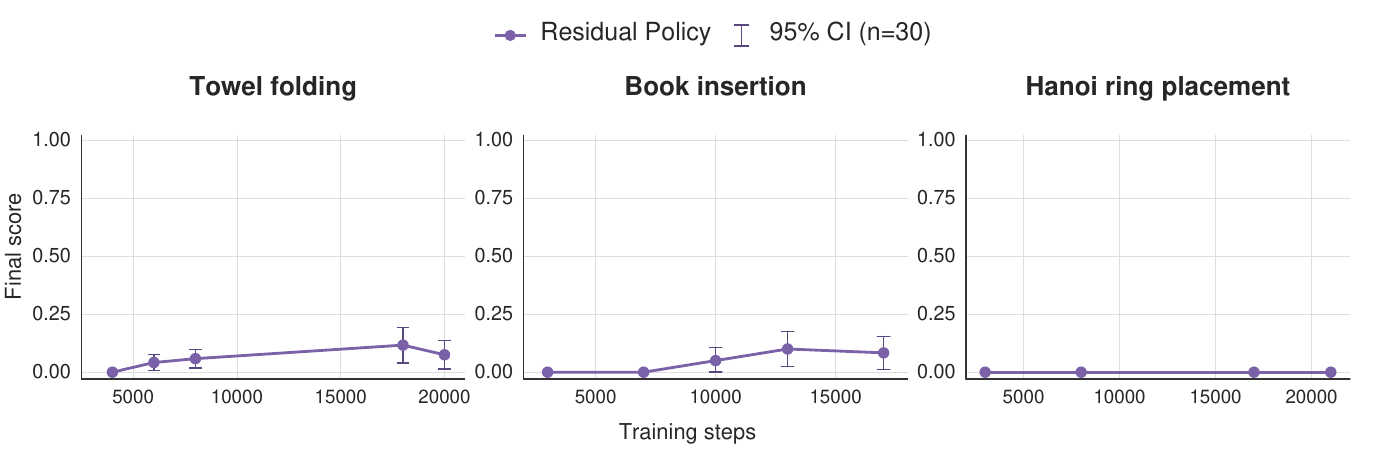}
  \vspace{-4pt}
  \caption{\textbf{Lightweight residual baseline on three tasks.} The lightweight residual module substantially underperforms LIFT.}
  \label{fig:residual}
\end{figure}

LIFT post-trains the full VLA network, which carries a substantial computational cost. This motivates a potentially cheaper alternative: can a dedicated lightweight residual network learn to react to force and correct the actions of a frozen base policy? We test this alternative with a force-reactive residual policy, but it achieves substantially lower scores than LIFT (\Cref{fig:residual}). Specifically, the weak base requires large corrections, while sparse residual targets make intervention timing difficult to learn. Implementation details are provided in \Cref{app:residual}.

\FloatBarrier
\section{Conclusion and Limitations}
\label{sec:conclusion}

LIFT adds force-aware post-training to pretrained VLA policies through late reactive force injection, prior-preserving initialization, and online corrective training. Across three tasks, it learns faster and reaches higher peak and final performance than vision-only DAgger while preserving the pretrained prior. This suggests that post-training can add a physically grounded modality to a pretrained VLA without disturbing its existing capabilities.

\textbf{Limitations and future directions.}
\label{sec:limitations}
Human intervention limits data-collection efficiency and places a substantial burden on operators, who must continuously decide whether to take over and remain ready to intervene. Takeover timing is difficult to judge, requires skilled operators, and can affect correction-data quality. Our experiments are conducted only on a single-arm platform. Future work should reduce the monitoring and intervention burden, study bimanual tasks on dual-arm platforms, and explore different force sensors.

\clearpage
\acknowledgments{We thank the reviewers for their constructive comments. We also thank Yuan Fang, Fangyuan Zhou, and Yanwen Zou for helpful discussions. We especially thank Shirun Tang for help with robot and controller debugging.

This work was supported by Fundamental and Interdisciplinary Disciplines Breakthrough Plan of the Ministry of Education of China, Science and Technology Major Project of Jiangsu Province (No.~SBG2024041), Shanghai Artificial Intelligence Laboratory, and XPLORER PRIZE grants.}


\bibliography{example}  


\ifdefined\LIFTNoAppendix
\else
\clearpage
\appendix
\section{Latency-aligned causal mask details}
\label{app:mask-details}

After the reactive action stream is made causal, LIFT injects force so each action can condition on the newest admissible force observations. The recent 6D end-effector force chunk $F_{t:t+H}\in\mathbb{R}^{H\times 6}$ is first encoded by a single-layer \textbf{causal} GRU followed by a linear projection,
\begin{equation}
\begin{aligned}
m_{t:t+H} \;=\; \mathrm{Linear}_{d_m}\!\big(\mathrm{GRU}_{d_h}(F_{t:t+H})\big) \;\in\; \mathbb{R}^{H\times d_m},
\label{eq:force}
\end{aligned}
\end{equation}
where $d_h=d_m/2$ and $d_m$ matches the action-expert width, and we denote the resulting sequence as the \emph{force memory} $m_{t:t+H}$.

Only the reactive action tokens attend to this force memory. Concretely, each reactive query $q_i$ attends over the full force memory sequence through force-injected cross attention, while a latency-aligned causal mask removes force tokens that should not yet be available:
\begin{equation}
\begin{aligned}
\mathrm{Cross}(q_i, m_{0:H-1})
&= \sum_{j=0}^{H-1} \alpha_{ij}\, m_j W_V,\\
\alpha_{ij}
&= \frac{\exp(s_{ij}+b^{(L)}_{ij})}
        {\sum_{k=0}^{H-1}\exp(s_{ik}+b^{(L)}_{ik})},\qquad
s_{ij}=\frac{(q_iW_Q)(m_jW_K)^\top}{\sqrt{d_m}},\\
b^{(L)}_{ij}
&=
\begin{cases}
0, & j \le i-L,\\
-\infty, & j > i-L .
\end{cases}
\end{aligned}
\label{eq:cross}
\end{equation}
Here $L$ is a latency-alignment offset chosen to match the network inference delay. The mask is causal in the force stream: it prevents the reactive action expert from using force measurements that should not yet be available and shifts the available force window to the action time after inference completes. The shift ensures that after a nonzero delay, the model still outputs the action to execute at completion time rather than an outdated action.

\section{Pseudocode: cached reactive inference}
\label{app:pseudo}
For completeness, \Cref{alg:reactive-inference} restates the cached reactive rollout loop referenced in \Cref{sec:method}.

\begin{algorithm}[H]
\caption{Cached reactive inference with latency-aligned force injection.}
\label{alg:reactive-inference}
\small
\begin{algorithmic}[1]
\Require policy $\pi_\theta$, language instruction $\ell$, image stream, force stream, execution horizon $H$, diffusion noise level $N$, latency offset $L$
\State $t \leftarrow 0$
\While{task not done}
  \State $I_t \leftarrow \mathrm{GetImage}()$
  \State $K_{\mathrm{VL}},V_{\mathrm{VL}}\leftarrow \mathrm{VLA}_{\mathrm{prefix}}(\ell,I_t)$ \Comment{cache slow vision-language context}
  \State $x_N, x_N^{\mathrm{r}} \sim \mathcal{N}(0,\mathbf{I})$ \Comment{initialize base and reactive action noise}
  \State cache the latency-aligned causal mask $b^{(L)}$ with $b^{(L)}_{ij}=-\infty$ for $j>i-L$
  \State $\mathcal{F}\leftarrow [\,]$ \Comment{clear force memory history for this chunk}
  \For{$i \leftarrow 0$ \textbf{to} $H-1$}
    \State $f_{t+i}\leftarrow \mathrm{GetForceFrame}()$
    \State append $f_{t+i}$ to $\mathcal{F}$
    \State $m_{t:t+i}\leftarrow \mathrm{ForceEncoder}(\mathcal{F})$ using \eqref{eq:force}
    \State $\hat{a}_{t+i}\leftarrow \mathrm{DenoiseAction}(\pi_\theta,K_{\mathrm{VL}},V_{\mathrm{VL}},x_N,x_N^{\mathrm{r}},m_{t:t+i},b^{(L)})$
    \State $\mathrm{Execute}(\hat{a}_{t+i})$
    \If{task done}
      \State \textbf{break}
    \EndIf
  \EndFor
  \State $t \leftarrow t+H$ \Comment{refresh slow context, noise, and force history}
\EndWhile
\end{algorithmic}
\end{algorithm}

\section{Online DAgger protocol}
\label{app:dagger}

\paragraph{Data sources.} Two datasets are mixed at training time: an offline visual dataset $\mathcal{D}_v$ (stage-one handheld), and a growing online dataset $\mathcal{D}_f$ (stage-two online correction data) that the collection pipeline appends as each new rollout terminates. The online data are force-enabled.

\paragraph{Fixed data mixing.} The main experiments use a fixed $1{:}1$ offline:online sampling ratio. Each training batch therefore contains equal amounts of offline task-alignment data and online force-enabled correction data.

\paragraph{Synchronization.} The trainer checks the data-cloud endpoint for newly completed episodes at every training step. Updated checkpoints are pushed to the inference server every $100$ training steps, which refreshes the policy deployed on the robot.

\paragraph{Human intervention.} During each rollout, an operator monitors the robot and triggers a reset whenever (i) the policy stalls in a non-progressing configuration, (ii) the force trace exceeds a hardware safety threshold, or (iii) the episode exceeds the task's max length.

\section{Hardware and sensor setup}
\label{app:hardware}

\paragraph{Robot.} A single Flexiv Rizon~4S 7-DoF cobot equipped with a Robotiq 2F-85 parallel-jaw gripper executes all rollouts. The arm operates in position-control mode at 10\,Hz, and trajectory targets from the policy are tracked by the robot's internal impedance controller.

\paragraph{Sensing.} A built-in 6D end-effector force/torque (F/T) sensor provides raw wrench measurements $f_x, f_y, f_z, \tau_x, \tau_y, \tau_z$ at more than 1000\,Hz. We downsample this stream to 10\,Hz and time-synchronize it with the action stream before using it as policy input. A wrist-mounted iPhone provides RGB images at 10\,Hz. We use a single-camera setup for all tasks reported here.

\paragraph{Teleoperation.} Stage-two force-feedback demonstrations are collected through Flexiv TDK~\citep{flexivtdk2026}. TDK uses a bilateral master-follower setup with two Flexiv Rizon~4S arms: the operator moves the master arm, while the follower arm executes the task. Contact forces measured on the follower arm are fed back to the master arm, allowing the operator to feel force signals and adjust the motion.

\section{Dataset details}
\label{app:dataset}

\paragraph{Stage-one offline data (stage-one handheld).} The stage-one offline dataset $D_v$ is collected with a handheld gripper with iPhone (no force sensor) at 10\,Hz. We use the iPhone main camera for RGB recording and call the ARKit SLAM interface to estimate the camera pose. Each demonstration is recorded and represented in the iPhone camera frame. We remove short static segments from the beginning and end of each episode to avoid stuck frames.

\paragraph{Stage-two online data (stage-two online correction data).} Collected with the Flexiv TDK setup. During collection, we set the robot TCP to the iPhone camera frame. Each timestep stores the synchronized 6D wrench alongside RGB observations. We record robot commands rather than robot states to provide more accurate control supervision. Only human-intervention correction data are added to the stage-two dataset $D_f$.

\paragraph{Action representation.} All actions are relative actions. The action vector is padded to the shared output dimension required by the policy head.

\clearpage
\section{Training hyperparameters}
\label{app:hparams}

\begin{table}[H]
\centering
\caption{\textbf{Architecture hyperparameters.}}
\label{tab:arch-hparams}
\small
\begin{tabular}{ll}
\toprule
Parameter & Value \\
\midrule
Backbone & $\pi_{0.5}$ with PaliGemma vision-language tower and action expert \\
Action chunk horizon & $10$ \\
Padded action dimension & $32$ \\
Force input dimension & $6$D wrench \\
Force encoder & Single-layer causal GRU followed by a linear projection \\
Force encoder hidden width & $512$ \\
Force encoder output width & $d_m=1024$ \\
Latency-alignment offset & $3$ \\
Reactive expert initialization & Copied from the base action expert \\
Cross-attention initialization & Zero-initialized output projection \\
\bottomrule
\end{tabular}
\end{table}

\begin{table}[H]
\centering
\caption{\textbf{Optimization hyperparameters.}}
\label{tab:opt-hparams}
\small
\setlength{\tabcolsep}{4pt}
\begin{tabular}{p{0.48\linewidth}p{0.36\linewidth}}
\toprule
Parameter & Value \\
\midrule
Batch size & $32$ \\
Total training steps & $30{,}000$ \\
Learning-rate schedule & $10{,}000$-step warmup, cosine schedule \\
Peak and final learning rate & $5\!\times\!10^{-5}$; decay horizon $1{,}000{,}000$ \\
Optimizer & AdamW \\
AdamW $\beta_1$ & $0.9$ \\
AdamW $\beta_2$ & $0.95$ \\
AdamW $\varepsilon$ & $10^{-8}$ \\
Weight decay & $10^{-10}$ \\
Gradient clipping norm & $1.0$ \\
\bottomrule
\end{tabular}
\end{table}

\section{Compute and reproducibility}
\label{app:compute}
\paragraph{Compute.} Each training run uses a single 8-GPU node equipped with NVIDIA A800 GPUs.

\paragraph{Code release.} Our codebase is built on \texttt{openpi} and is publicly available at \url{https://github.com/y-wng/lift}.

\section{Residual baseline implementation}
\label{app:residual}

Our custom baseline is inspired by ImplicitRDP~\citep{chen2025implicitrdp} and CR-DAgger~\citep{xu2025crdagger}, not a reproduction of either system. It freezes the offline-trained $\pi_{0.5}$ policy and trains an approximately 28M-parameter causal Transformer with a force GRU to predict additive position-action corrections. Visual features are cached at 1\,Hz and corrections are produced at 10\,Hz. Training uses a fixed $1{:}1$ offline:online mixture and mean squared error on scaled residual targets: expert minus base actions during intervention, and zero otherwise.

\clearpage
\section{Task protocol and failure case analysis}
\label{app:tasks}
\label{app:failures}
Each rollout lasts at most $2$\,min. For towel folding, the robot starts $5\,\text{cm}$ directly above the lowest towel corner, with the towel reset before each rollout. For Hanoi, the robot starts holding the ring above the base of a $10\,\text{mm}$-diameter pole. The qualitative examples are shown in \Cref{fig:task-protocol-failures}.

\begin{figure}[H]
  \centering
  \includegraphics[width=\linewidth]{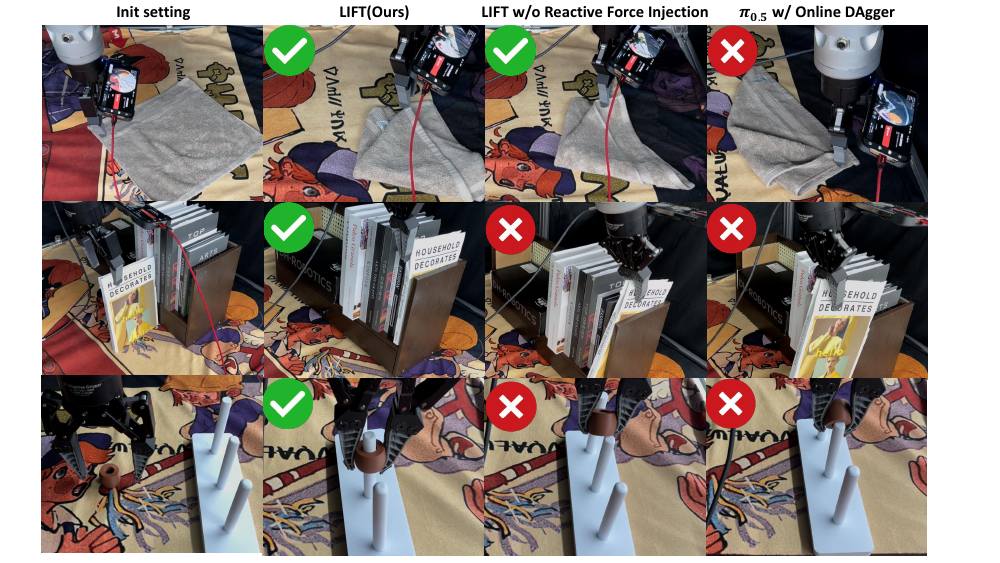}
  \vspace{-4pt}
  \caption{\textbf{Task protocol and failure case analysis.} Qualitative examples of towel folding, book insertion, and Hanoi ring placement.}
  \label{fig:task-protocol-failures}
  \label{fig:qualitative-cases}
\end{figure}
\paragraph{\textbf{Towel folding.}}
The towel score is determined by the highest completed stage: grasping the first corner gives $0.25$, completing the first fold gives $0.5$, grasping the second corner gives $0.75$, and completing the second fold gives $1.0$. Common failures include an empty grasp, grasping the wrong second corner, and an incorrect folding motion.

\paragraph{\textbf{Book insertion.}}
The robot receives $0.5$ for grasping and lifting the book and inserting it without collision. It receives $1.0$ after closing the gripper, pushing the spine until the book is fully inserted, and disengaging from contact. Common failures include side collision or early release, over-pushing or missing the spine, incomplete insertion, and failure to disengage.

\paragraph{\textbf{Hanoi ring placement.}}
The Hanoi task is binary: a ring fully seated on the pole base receives $1$, and an unsuccessful placement receives $0$. Common failures include angled insertion and jamming, as well as impact-induced oscillation or drift.
\clearpage
\section{Training-step view of learning curves}
\label{app:steps-curves}

Training steps count optimizer updates and are positively correlated with training time. Each online experiment takes approximately 2--3 hours and collects 20--30 episodes. These episode counts describe online collection; only human-intervention frames are used for online training, as described in the experimental setup.

\begin{figure}[H]
  \centering
  \includegraphics[width=\linewidth]{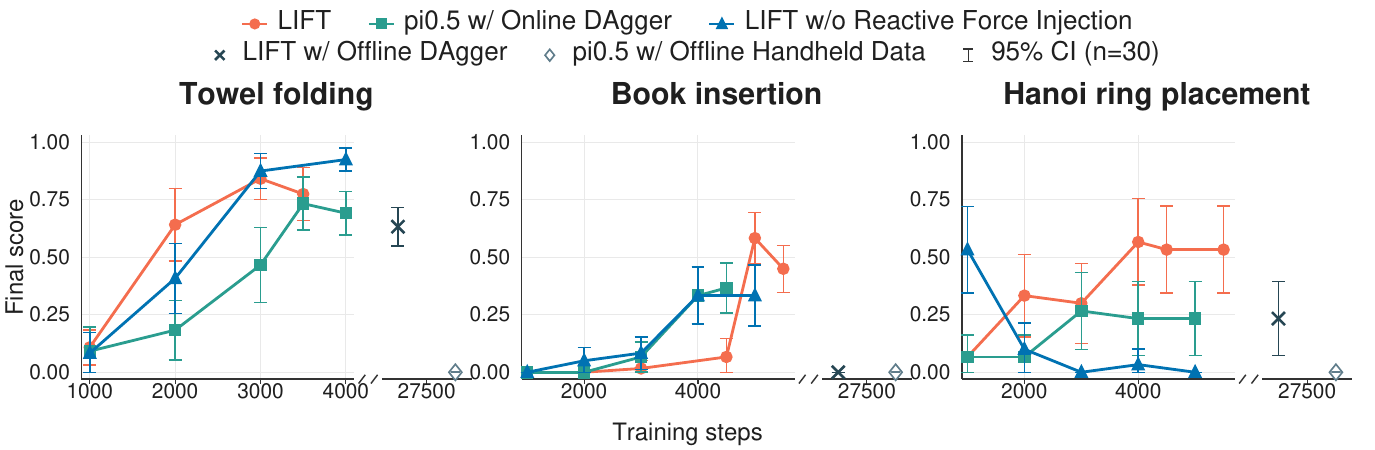}
  \vspace{-4pt}
  \caption{\textbf{Learning curves on three tasks as optimizer training steps increase.}
  The horizontal axis counts optimizer updates; training steps are positively correlated with training time.}
  \label{fig:online-curves-steps}
\end{figure}

\FloatBarrier

\section{Additional ablations}
\label{app:ablations}

\subsection{How does the offline:online data ratio affect post-training? (Q6)}
\label{sec:q-offline}
We vary the offline:online sampling ratio on towel folding to examine its effect on post-training. The $0{:}1$ ratio performs substantially worse than $1{:}1$ and $1{:}2$, with $1{:}1$ improving faster (\Cref{fig:ratio}). All main experiments use fixed $1{:}1$ offline:online sampling.

\begin{figure}[H]
  \centering
  \includegraphics[width=0.64\linewidth]{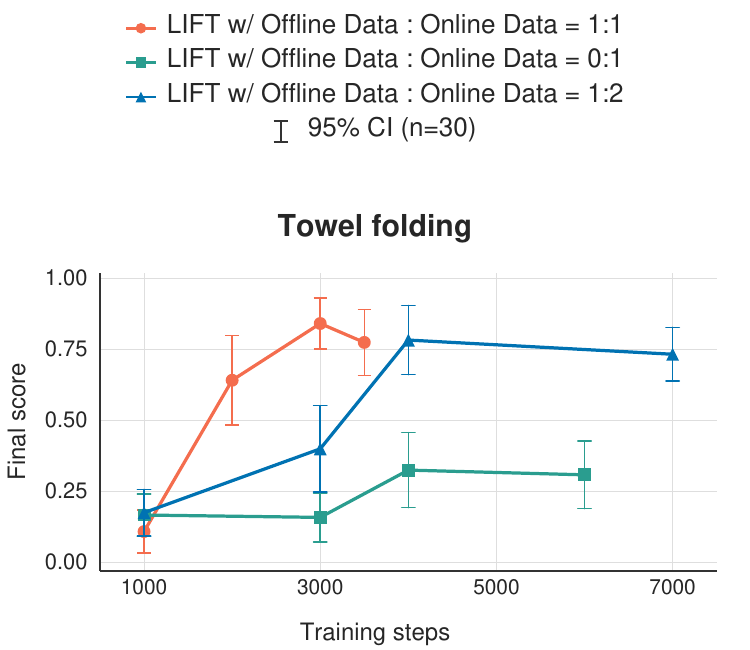}
  \vspace{-2pt}
  \caption{\textbf{Offline data ratio ablation.} Mixed training outperforms online-only training; the main protocol uses fixed $1{:}1$ sampling.}
  \label{fig:ratio}
\end{figure}

\clearpage
\section{Generalization condition details}
\label{app:gen}

\begin{figure}[H]
  \centering
  \includegraphics[width=\linewidth]{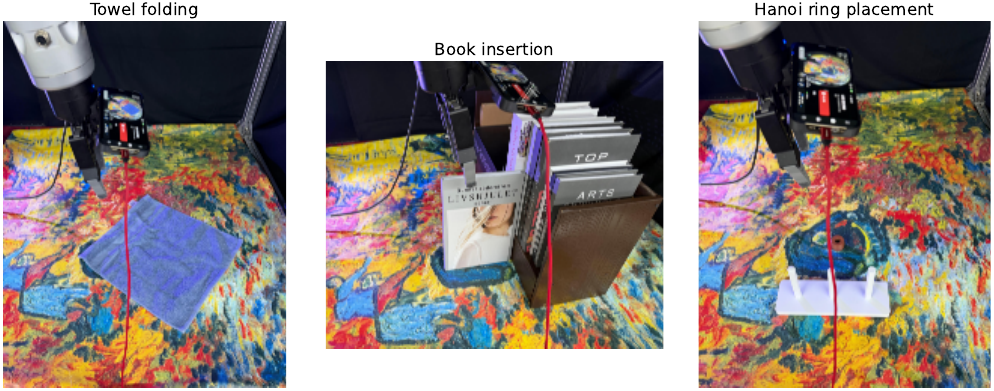}
  \vspace{-0.5em}
  \caption{\textbf{Generalization task settings.} The three panels show the towel folding, book insertion, and Hanoi ring placement setups used for shifted-condition evaluation.}
  \label{fig:gen-settings}
\end{figure}

\paragraph{Towel folding.}
\begin{itemize}[leftmargin=2.0em,labelsep=0.5em,itemsep=2pt]
  \item \emph{Blue towel:} replace the original towel with a blue towel of the same style.
  \item \emph{Another tablecloth:} replace the rough cartoon-style brown tablecloth with a slightly smoother yellow tablecloth with an oil-painting texture.
  \item \emph{Purple light:} add a purple side light while keeping the task geometry unchanged.
\end{itemize}

\paragraph{Book insertion.}
\begin{itemize}[leftmargin=2.0em,labelsep=0.5em,itemsep=2pt]
  \item \emph{Gray book:} replace the original book with a new gray book.
  \item \emph{Another tablecloth:} replace the rough cartoon-style brown tablecloth with the same smoother yellow oil-painting-texture tablecloth used in the towel-folding shift.
  \item \emph{Purple light:} add the same purple side light.
\end{itemize}

\paragraph{Hanoi ring placement.}
\begin{itemize}[leftmargin=2.0em,labelsep=0.5em,itemsep=2pt]
  \item \emph{Another tablecloth:} replace the rough cartoon-style brown tablecloth with the smoother yellow oil-painting-texture tablecloth.
  \item \emph{Purple light:} add the same purple side light.
\end{itemize}

For every shift the robot start pose and target object location are kept identical to the in-distribution evaluation; only the listed factor changes.


\fi
\end{document}